\documentclass[10pt,twocolumn,letterpaper]{article}

\usepackage{iccv}
\usepackage{times}
\usepackage{epsfig}
\usepackage{graphicx}
\usepackage{amsmath}
\usepackage{amssymb}

\usepackage{makecell}
\usepackage{multirow}
\usepackage{multicol}
\usepackage{subfig}
\usepackage{lipsum}
\usepackage{xcolor}
\usepackage{dsfont}
\usepackage{pifont}
\usepackage{booktabs}

\usepackage[breaklinks=true,bookmarks=false,colorlinks]{hyperref}

\iccvfinalcopy

\def\iccvPaperID{2833}

\begin{document}

\title{Beyond Human Parts: Dual Part-Aligned Representations\\ for Person Re-Identification}

\author{
    Jianyuan Guo$^{1\dagger}$, Yuhui Yuan$^{2,3\dagger}$,
    Lang Huang$^1$, Chao Zhang$^{1}$\thanks{Corresponding author. $^\dagger$Equal contribution.}, Kai Han, Jin-Ge Yao\\
    $^1$Key Laboratory of Machine Perception (MOE), Peking University \\
    $^2$Institute of Computing Technology \;
    $^3$University of Chinese Academy of Sciences \\
    {\tt \small \{jyguo, laynehuang\}@pku.edu.cn, chzhang@cis.pku.edu.cn}\\ 
    {\tt \small \{yuhui.yuan, jinge.yao\}@microsoft.com, kai.han@huawei.com}}

\maketitle
\begin{abstract}
\vspace{-0.3cm}
Person re-identification is a challenging task due to various complex factors. 
Recent studies have attempted to integrate human parsing results or externally defined attributes to help capture human parts or important object regions.
On the other hand, there still exist many useful contextual cues that do not fall into the scope of predefined human parts or attributes.
In this paper, we address the missed contextual cues by exploiting
both the accurate human parts and the coarse non-human parts.
In our implementation, we apply a human parsing model to extract
the binary human part masks \emph{and} a self-attention mechanism
to capture the soft latent (non-human) part masks.
We verify the effectiveness of our approach with new state-of-the-art performances on three challenging benchmarks: Market-$1501$, DukeMTMC-reID and CUHK$03$. Our implementation is available at \href{https://github.com/ggjy/P2Net.pytorch}{https://github.com/ggjy/P2Net.pytorch}.

\end{abstract}
\vspace{-0.5cm}
\section{Introduction}
\vspace{-0.2cm}
Person re-identification has attracted increasing attention from both the academia and the industry in the past decade due to its significant role in video surveillance. 
Given an image for a particular person captured by one camera, the goal is to re-identify this person from images captured by different cameras from various viewpoints. 

The task of person re-identification is inherently challenging because of the significant visual appearance changes caused by various factors such as human pose variations, lighting conditions, part occlusions, background cluttering and distinct camera viewpoints. 
All these factors make the misalignment problem become
one of the most important problems in person re-identification task.
With the surge of interest in deep representation learning, various approaches
have been developed to address the misalignment problem, 
which could be roughly summarized as the following streams:
(1) Hand-crafted partitioning, which relies on manually designed splits such as grid cells \cite{deepreid,nips_16_rigid_part,patch} or horizontal stripes \cite{ahmed2015improved,cheng2016person,pcb,varior2016siamese,yi2014deep} of the input image or feature maps, based on the assumption that human parts are well-aligned in the RGB color space.
(2) The attention mechanism, which tries to learn an attention map over the last output feature map and constructs the aligned part features accordingly \cite{zhaoliming,cvpr_18_dual_attention_part, aacn, wang2018mancs}. 
(3) Predicting a set of predefined attributes \cite{layne2014attributes,su2016attributes,lin2017improving,chang2018MLFN,su2018multi} as useful features to guide the matching process. 
(4) Injecting human pose estimation \cite{cvpr_16_pose_aware,cvpr_17_pose_person_recognition,cvpr_18_pose_transfer,pdc,aacn,spindlenet,eccv_18_pose_norm} or human parsing result \cite{spreid,lianglip,mgcam} to extract the human part aligned features based on the predicted human key points or semantic human part regions, while the success of such approaches heavily count on the accuracy of human parsing models or pose estimators.
Most of the previous studies mainly focus on learning more accurate human part representations,
while neglecting the influence of potentially useful contextual cues that could be addressed as ``non-human'' parts.

\begin{figure*}[!t]
\centering
\includegraphics[width=\linewidth]{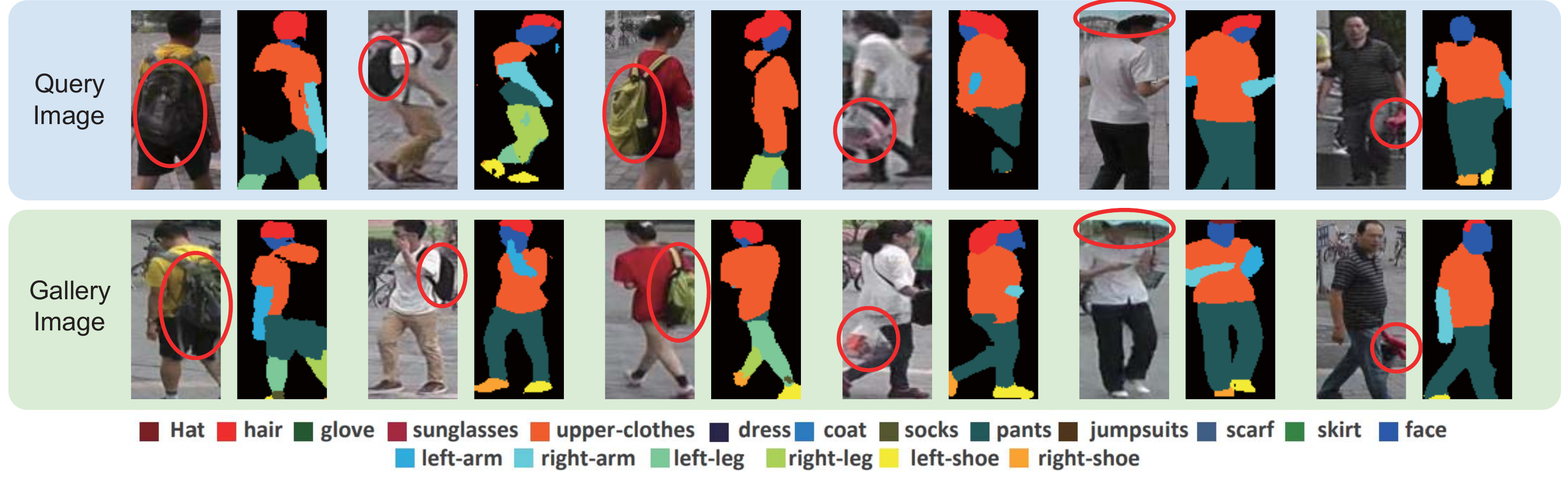}
\vspace{-0.7cm}
\caption{\small{\textbf{Failure cases of the human parsing model}: 
The first row illustrates the query images,
the second row illustrates the gallery images from Market-$1501$
and each column consists of two images belonging to the same identity.
All of the regions marked with red circle are mis-classified as background (marked with black color)
due to the limited label set
while 
their ground-truth labels should be backpack, reticule and umbrella.
It can be seen that these mis-classified regions
are crucial for the person re-identification.
}}
\label{fig:intro}
\vspace{-0.4cm}
\end{figure*}

Existing human parsing based approaches~\cite{aacn,spindlenet} utilize an off-the-shelf semantic segmentation model to divide the input image into $K$ predefined human parts, according to a predefined label set.\footnote{\Eg the label set in \cite{lianglip}: background, hat, hair, glove, sunglasses, upper-clothes, dress,
coat, socks, pants, jumpsuits, scarf, skirt, face, right-arm, left-arm,
right-leg, left-leg, right-shoe and left-shoe.}
Beyond these predefined part categories,
there still exist many objects or parts which could be critical for person re-identification, but tend to be recognized as background by the pre-trained human parsing models. 
For example, we illustrate some failure cases
from human parsing results on the Market-$1501$ dataset in Figure~\ref{fig:intro}.
We can find that the
objects belonging to undefined categories such as 
\textbf{backpack, reticule} and \textbf{umbrella}
are in fact helpful and sometimes crucial for person re-identification.
The existing human parsing datasets are mainly
focused on parsing human regions, 
and most of these datasets fail to include all possible identifiable objects that could help person re-identification.
Especially, most of the previous attention based approaches are mainly focused on extracting the human part attention maps. 

Explicitly capturing useful information beyond predefined human parts or attributes has not been well studied in the previous literature.
Inspired by the recently popular self-attention scheme~\cite{attentionisallyouneed,nonlocal}, we attempt to address the above problem by learning latent part masks from the raw data,
according to the appearance similarities among pixels, which provide a coarse estimation of
both human parts and the non-human parts,
with the latter largely overlooked from the previous approaches based on human parsing.

Moreover, we propose a \textbf{dual part-aligned representation} scheme 
to combine the complementary information from 
both the accurate human parts and 
the coarse non-human parts.
In our implementation, 
we apply a human parsing model to extract the human part masks 
and compute the human part-aligned representations for the features
from low-levels to high-levels.
For the non-human part information,
we apply self-attention mechanism to learn
to group all pixels belonging to the same latent part together.
We also extract the latent non-human part information
on the feature maps
from the low-levels to the high-levels.
Through combining the advantages of both the accurate human part information
and the coarse non-human part information, our approach learns to 
augment the representation of each pixel with the representation
of the part (human parts \emph{or} non-human parts) that it belongs to.

Our main contributions are summarized as below:
\begin{itemize}
\vspace{-0.28cm}
\item We propose the \textbf{dual part-aligned representation} to update the representation
by exploiting the complementary information from both the accurate human parts and the coarse non-human parts.
\vspace{-0.28cm}
\item We introduce the $P^2$-Net and show that our $P^2$-Net achieves new state-of-the-art performance on three benchmarks including Market-$1501$, DukeMTMC-reID and CUHK$03$.
\vspace{-0.28cm}
\item We analyze the contributions from both the human part representation and the latent part (non-human part) representation and discuss their complementary strengths in our ablation studies.
\end{itemize}

\section{Related Work}
\vspace{-0.2cm}

The part misalignment problem is one of the key challenges for person re-identification, a host of methods \cite{zhaoliming, pdc, li2017learning, spindlenet, pcb, eccv_18_pose_norm, cvpr_17_pose_person_recognition, spreid, mgcam, cvpr_18_part_conv, aacn, nips_16_rigid_part, cvpr_18_dual_attention_part, cvpr_18_rigid_part} 
have been proposed to mainly exploit the human parts to handle the body part misalignment problem, we briefly summarize the existing methods as below:

\vspace{-0.5cm}
\paragraph{Hand-crafted Splitting for ReID.}
In previous studies, there are methods proposed to divide the input image or the feature map into small patches \cite{ahmed2015improved,deepreid,nips_16_rigid_part} or stripes \cite{cheng2016person,varior2016siamese,yi2014deep} and then extract region features from the local patches or stripes. 
For instance, PCB~\cite{pcb} adopts a uniform partition and 
further refines every stripe with a novel mechanism.
The hand-crafted approaches depend on the strong assumption that the spatial distributions of human bodies and human poses are exactly matching.

\vspace{-0.5cm}
\paragraph{Semantic Segmentation for ReID.}
Different from the hand-crafted splitting approaches, \cite{pse,pdc,spindlenet,spreid} apply a human part detector or a human parsing model to capture more accurate human parts. For example, SPReID \cite{spreid} utilizes a parsing model to generate $5$ different predefined human part masks to compute more reliable part representations, which achieves promising results on various person re-identification benchmarks.

\vspace{-0.5cm}
\paragraph{Poses/Keypoints for ReID.}
Similar to the semantic segmentation approaches, poses or keypoints estimation can also be used for accurate/reliable human part localization.
For example, 
there are approaches exploring both the human poses and the human part masks~\cite{cvpr_14_pose},
or generating human part masks via exploting the connectivity of the keypoints~\cite{aacn}.
There are some other studies \cite{cvpr_16_pose_aware,pse,pdc,spindlenet} that also exploit the pose cues to extract the part-aligned features.

\vspace{-0.5cm}
\paragraph{Attention for ReID.}
Attention mechanisms have been used to capture human part information in recent work~\cite{liu2017end,zhaoliming,aacn,hacnn,mgcam}.
Typically, the predicted attention maps distribute most of the attention weights on human parts that may help improve the results. 
To the best of our knowledge, 
we find that most of the previous attention approaches are limited to capturing the human parts only.

\vspace{-0.5cm}
\paragraph{Attributes for ReID.}
Semantic attributes \cite{iccv_17_attribute_recognition,iccv_17_pedestrian_analysis,AAP} have been exploited as feature representations for person re-identification tasks.
Previous work
\cite{cvpr_18_attribute_identity, han2018attribute, lin2017improving,tay2019aanet,zhao2019attribute} leverages the attribute labels provided by original dataset to generate attribute-aware feature representation. 
Different from previous work, our latent part branch can attend to important visual cues without relying on detailed supervision signals from the limited predefined attributes.

\vspace{-0.5cm}
\paragraph{Our Approach.}

To the best of our knowledge, we are the first to explore and define the (non-human) contextual cues.
We empirically demonstrate the effectiveness of combining
separately crafted components for the well-defined, accurate human parts \emph{and} all other potentially useful (but coarse) contextual regions. 

\begin{figure*}[!t]
\centering
\includegraphics[width=\linewidth]{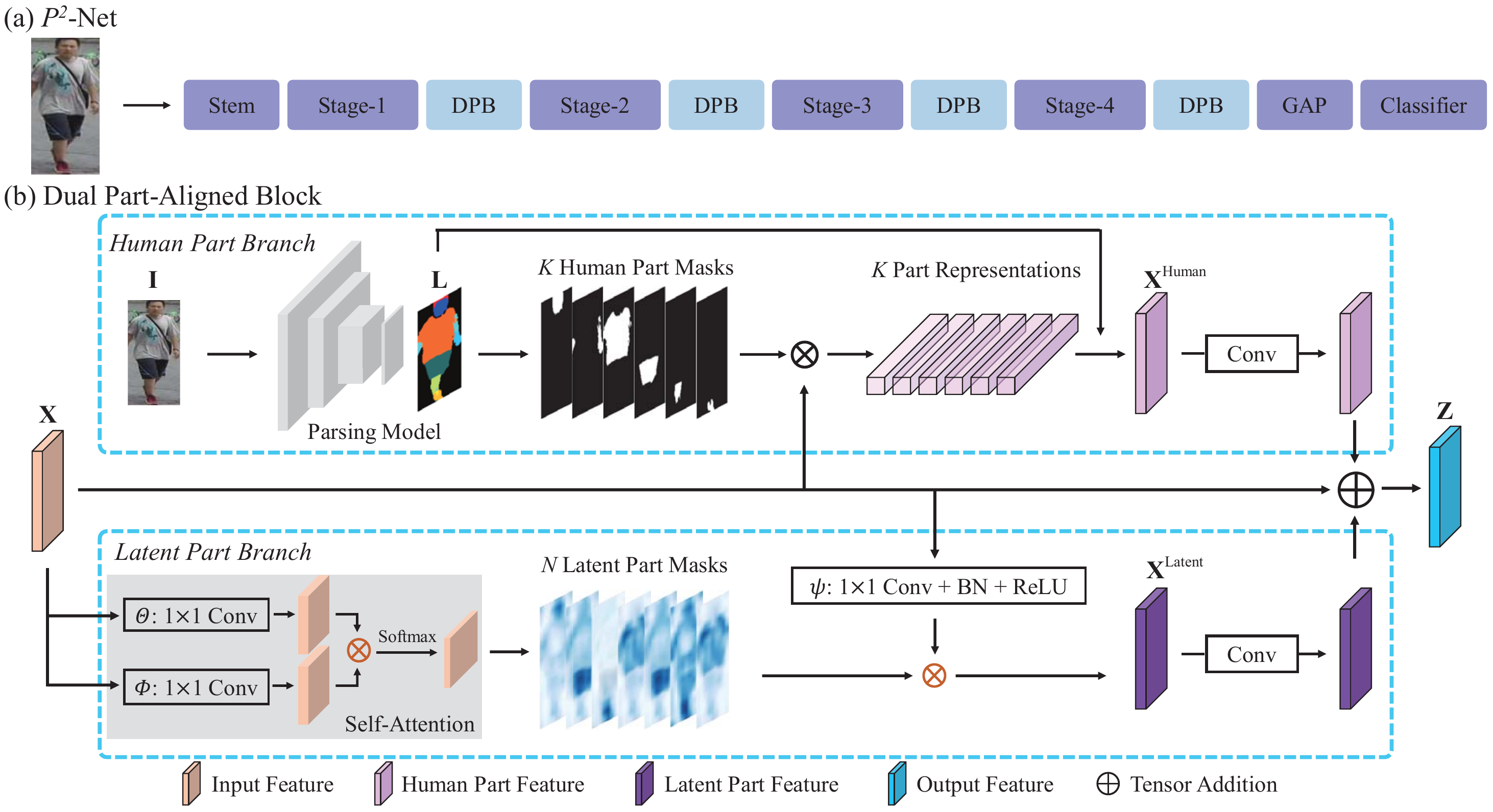}
\vspace{-0.5cm}
\caption{\small{Illustration of the overall structure of $P^2$-Net and the dual part-aligned block (DPB).
(a) Given an input image, we employ a ResNet-50 backbone consists of a stem, 
four stages (e.g., Res-1, Res-2, Res-3 and Res-4), 
global average pooling (GAP) and a classifier.
We insert a DPB after every stage within the ResNet backbone.
(b) The DPB consists of a human part branch and a latent part branch.
For the human part branch,
we employ the CE2P \cite{liu2018devil} to predict the human part label maps
and generate the human part masks accordingly.
For the latent part branch, we employ the self-attention scheme to predict the latent part masks. 
We compute the human part-aligned representation
and latent part-aligned representation within the two branches separately.
Last,
we add the outputs from these two branches to the input feature map as the final output feature map.
}}
\label{fig:hppnet}
\vspace{-0.3cm}
\end{figure*}

\vspace{-0.2cm}
\section{Approach}
First, we present our key contribution: \textbf{dual part-aligned representation}, which learns to combine both the accurate human part
information and the coarse latent part information
to augment the representation of each pixel (Sec.~\ref{approach2}).
Second, we present the network architecture and the detailed implementation of $P^2$-Net (Sec.~\ref{approach3}).

\subsection{Dual Part-Aligned Representation} \label{approach2}
Our approach consists of two branches: a \emph{human part branch} and a \emph{latent part branch}.
Given an input feature map $\mathbf{X}$ of size $N\times C$, where $N=H\times W$, $H$ and $W$ are the height and width of the feature map, 
$C$ is the number of channels,
we apply the human part branch to extract accurate human part masks
and compute the human part-aligned representation $\mathbf{X}^{\mathrm{Human}}$ accordingly.
We also use a latent part branch to learn to capture both the coarse
non-human part masks and the coarse human part masks based on the appearance similarities between different pixels, then we compute the latent part-aligned representation 
$\mathbf{X}^{\mathrm{Latent}}$ according to the coarse part masks.
Last, we augment the original representation with both the human part-aligned representation and the latent part-aligned representation.

\noindent\textbf{Human Part-Aligned Representation.}
The main idea of the human part-aligned representation is to represent each pixel with the human part representation that the pixel belongs to, which is the aggregation of the pixel-wise representations weighted by a set of confidence maps.
Each confidence map is used to surrogate a semantic human part.

We illustrate how to compute the human part-aligned representation in this section. 
Assuming there are $K-1$ predefined human part categories in total from a human parsing model,
we treat all the rest proportion of regions in the image
as background according to the human parsing result. 
In summary, we need to estimate $K$ confidence maps for the human part branch.

We apply the state-of-the-art human parsing framework CE2P~\cite{liu2018devil} to predict the semantic human part masks for all the images in all three benchmarks in advance, as shown in Figure~\ref{fig:hppnet}(b).
We denote the predicted label map of image $\mathbf{I}$ as $\mathbf{L}$. We re-scale the label map $\mathbf{L}$ to be of the same size as the feature map $\mathbf{X}$
($\mathbf{x}_i$ is the representation of pixel $i$, essentially the $i_{th}$ row of $\mathbf{X}$) before using it.
We use $l_i$ to represent the human part category of
pixel $i$ within the re-scaled label map, and $l_i$ is of $K$ different values
including $K-1$ human part categories and one background category.

We denote the $K$ confidence maps as ${\mathbf{P}_1,\mathbf{P}_2,\cdots,\mathbf{P}_K}$, where
each confidence map $\mathbf{P}_k$ is associated with a human part category (or the background category).
According to the predicted label map $\mathbf{L}$,
we set $p_{ki} = 1$ ($p_{ki}$ is the $i_{th}$ element of $\mathbf{P}_k$)
if $l_i \equiv k$ and $p_{ki} = 0$ otherwise.
Then we apply $\mathrm{L}1$ normalization on each confidence map and compute the human part representation as below,
\begin{align}
\textbf{h}_{k} &= g(\sum_{i=1}^{N} \hat{p}_{ki} \mathbf{x}_i), \label{eq:eq2}
\end{align}
where $\textbf{h}_{k}$ is the representation of the $k_{th}$ human part,
$g$ function is used to learn better representation
and $\hat{p}_{ki}$ is the confidence score after $\mathrm{L}1$ normalization.
Then we generate the human part-aligned feature map 
$\mathbf{X}^{\mathrm{Human}}$ of the same size as the
input feature map $\mathbf{X}$,
and each element of $\mathbf{X}^{\mathrm{Human}}$ is set as 
\begin{align}
\mathbf{x}^{\mathrm{Human}}_i &= \sum_{k=1}^{K} \mathds{1}[l_i\equiv k] \mathbf{h}_k, \label{eq:eq3}
\end{align}
where $\mathds{1}[l_i\equiv k]$ is an indicator function and 
each $\mathbf{x}^{\mathrm{Human}}_i$ is essentially the part representation of 
the semantic human part that it belongs to.
For the pixels predicted as the background, we
choose to aggregate
the representations of all pixels that are predicted as the background
and use it to augment their original representations.

\noindent\textbf{Latent Part-Aligned Representation.}
We explain how to estimate the latent part representation in this section. 
Since we can not predict accurate masks for non-human cues based on the existing approaches, we adapt the self-attention mechanism \cite{attentionisallyouneed,nonlocal}
to enhance our framework by learning to capture some coarse latent parts 
automatically from data based on the semantic similarities between
each pixel and other pixels.
The latent part is expected to capture details that are weakly utilized in the human part branch.
We are particularly interested in the contribution from
the coarse non-human part masks on the important cues that are missed by the predefined human parts or attributes.

In our implementation, the latent part branch learns to predict $N$ coarse confidence maps ${\mathbf{Q}_1,\mathbf{Q}_2,\cdots,\mathbf{Q}_N}$
for all $N$ pixels, each confidence map $\mathbf{Q}_i$ learns to pay
more attention to the pixels that belong to the same latent part
category as the $i_{th}$ pixel.

We illustrate how to compute the confidence map for
the pixel $i$ as below,
\begin{align}
q_{ij} = \frac{1}{Z_i} \exp(\theta (\mathbf{x}_j)^\top \phi(\mathbf{x}_i)), \label{eq:eq3}
\end{align}
where $q_{ij}$ is the $j_{th}$ element of $\mathbf{Q}_i$,
$\mathbf{x}_i$ and $\mathbf{x}_j$ are the representations of the pixels $i$ and $j$ respectively. 
$\theta(\cdot)$ and $\phi(\cdot)$ are two transform functions
to learn better similarities and are implemented as $1\times1$ convolution,
following the self-attention mechanism \cite{attentionisallyouneed,nonlocal}.
The normalization factor $Z_i$ is a sum of all the similarities
associated with pixel $i$:
$Z_i = \sum_{j=1}^N \exp(\theta(\mathbf{x}_j)^\top \phi(\mathbf{x}_i))$.

Then we estimate the latent part-aligned feature map $\mathbf{X}^{\mathrm{Latent}}$
as below,
\begin{align}
\mathbf{x}^{\mathrm{Latent}}_i &= \sum_{j=1}^{N} q_{ij} \psi(\mathbf{x}_j), \label{eq:eq4}
\end{align}
where $\mathbf{x}^{\mathrm{Latent}}_i$ is the $i_{th}$
element of $\mathbf{X}^{\mathrm{Latent}}$.
We estimate the latent part-aligned representation
for pixel $i$ by aggregating the representations of all the other
pixels according to their similarities with pixel $i$.
$\psi$ is a function used to learn better representation,
which is implemented with $1\times1$ convolution + BN + ReLU.

For the latent part-aligned representation, 
we expect each pixel can pay more attention to the part
that it belongs to, which is similar with the recent work~\cite{OCNet,huang2019isa,yuan2019ocr}.
The self-attention is a suitable mechanism to 
group the pixels with similar appearance together.
We empirically study the influence of the coarse human part information
and the coarse non-human part information
to verify the effectiveness is mainly attributed to the coarse non-human parts (Sec.~\ref{ablation-study}).

Last,
we fuse the human part-aligned representation
and the latent part-aligned representation as below,
\begin{align}
\textbf{Z} &= \textbf{X} + \mathbf{X}^{\mathrm{Human}} + \mathbf{X}^{\mathrm{Latent}},
\label{eq:merge}
\end{align}
where $\textbf{Z}$ is the final representation of our DPB block.

\subsection{$P^2$-Net}
\noindent\textbf{Backbone}.
We use ResNet-$50$ pre-trained on ImageNet as the backbone following the previous PCB~\cite{pcb}.

\noindent\textbf{Dual Part-Aligned Representation}.
In our implementation, we employ the dual part-aligned block (DPB) after Res-1, Res-2, Res-3 and Res-4 stages. 
Assuming that the input image is of size $384 \times 128$, 
the output feature map from Res-1/Res-2/Res-3/Res-4 stage is of size $96 \times 32$/$48 \times 16$/ $24 \times 8$/$24 \times 8$ respectively.
We have conducted detailed ablation study about DPB in Section~\ref{ablation-study}.
For the human part branch, we employ the CE2P~\cite{liu2018devil} model to extract the human part label maps of size $128\times64$, then we resize the label maps to be of the size $96 \times 32$/$48 \times 16$/$24 \times 8$/$24 \times 8$ for the four stages respectively.
For the latent part branch, we employ the self-attention mechanism on the output feature map
from each stage directly.

\noindent\textbf{Network Architecture}. \label{approach3}
The ResNet backbone takes an image $\mathbf{I}$ as input and outputs feature map $\mathbf{X}$ after the Res-4 stage. We feed the feature map $\mathbf{X}$ into the global average pooling layer and employ the classifier at last.
We insert the DPB after every stage to update the representation before feeding the feature map into the next stage. 
We could achieve better performance
through applying more DPBs.
The overall pipeline is illustrated in Figure~\ref{fig:hppnet}(a).

\noindent\textbf{Loss Function}.
All of our baseline experiments only employ the softmax loss to 
ensure the fairness of the comparison and for ease of ablation study.
To compare with the state-of-the-art approaches, 
we further employ the triplet loss (details in appendix)
following the previous work.

\vspace{-0.1cm}
\section{Experiments}

\subsection{Datasets and Metrics}

\vspace{-0.1cm}
\paragraph{Market-1501.} Market-$1501$ dataset \cite{market1501} consists of $1501$ identities captured by $6$ cameras, where the train set consists of $12,936$ images of $751$ identities, the test set is divided into a query set that contains $3,368$ images and a gallery set that contains $16,364$ images.  

\vspace{-0.45cm}
\paragraph{DukeMTMC-reID.} DukeMTMC-reID dataset \cite{duke,dukeprotocol} consists of $36,411$ images of $1,404$ identities captured by $8$ cameras, where the train set contains $16,522$ images, the query set consists of $2,228$ images and the gallery set consists of $17,661$ images.

\vspace{-0.45cm}
\paragraph{CUHK03.} CUHK03 dataset \cite{deepreid} contains $14,096$ images of $1,467$ identities captured by $6$ cameras. 
CUHK03 provides two types of data, hand-labeled (``labeled'') and DPM-detected (``detected'') bounding boxes, the latter type is more challenging due to severe bounding box misalignment and cluttered background.
We conduct experiments on both ``labeled" and ``detected" types of data. 
We split the dataset following the training/testing split protocol proposed in \cite{reranking}, where the train/query/gallery set consists of $7,368$/$1,400$/$5,328$ images respectively.

We employ two kinds of evaluation metrics including the cumulative matching characteristics (CMC) and mean average precision (mAP). 
Especially, all of our experiments employ the single-query setting without any other post-processing techniques such as re-ranking \cite{reranking}.

\vspace{-0.15cm}
\subsection{Implementation Details}
\vspace{-0.15cm}
We choose ResNet-50 pre-trained on ImageNet as our backbone. After getting the feature map from the last residual block, we use a global average pooling and a linear layer (FC+BN+ReLU) to compute a $256$-D feature embedding. 
We use ResNet-50 trained with softmax loss as our baseline model,
and set the stride of the last stage in ResNet from 2 to 1 following \cite{pcb}. 
We also use triplet loss \cite{cheng2016person,3d_nl_liao_video,zhaoliming} to improve the performance.

We use the state-of-the-art human parsing model CE2P \cite{liu2018devil} to predict 
the human part label maps for all the images in the three benchmark in advance. 
The CE2P model is trained on the Look Into Person \cite{lianglip} (LIP) dataset, which consists of $\scriptsize{\sim}30,000$ finely annotated images with 20 semantic labels (19 human parts and 1 background). 
We divide the 20 semantic categories into $K$ groups 
\footnote{When $K$ = 5, each group represents background, head, upper-torso, lower-torso and shoe; when $K$ = 2, it represents background and foreground;
when $K=1$, it treats the whole image as a single part.},
and train the CE2P model with the grouped labels. 
We adopt the training strategies as described in CE2P \cite{liu2018devil}.

All of our implementations are based on PyTorch framework \cite{paszke2017automatic}. 
We resize all the training images to $384\times128$ and 
then augment them by horizontal flip and random erasing \cite{randomerasing}.
We set the batch size as 64 and train the model with base learning rate starts from 0.05 and decays to 0.005 after 40 epochs, the training is finished at 60 epochs. 
We set momentum $\mu=0.9$ and the weight decay as $0.0005$. 
All of the experiments are conducted on a single NVIDIA TITAN XP GPU.

\begin{table*}[!ht]
\renewcommand\arraystretch{1.0}
\footnotesize
\centering
\caption{\small{
Ablation study of the DPB on Market-$1501$. 
$K$ is the number of human parts within the human part branch, 
We insert the DPB after the stage-$k$ (Res-$k$), where $k=1,2,3,4$. 
We employ HP-$p$ to represent the human part branch choosing $K=p$. 
DPB (HP-${p}$) represents using the human part branch only while 
DPB (Latent) represents using the latent part branch only.}}
\vspace{-0.35cm}
\label{table:res-k}
\begin{tabular}{l|ccc|ccc|ccc|ccc}
\hline
\multirow{2}{*}{Method} & \multicolumn{3}{c|}{Res-$1$} & \multicolumn{3}{c|}{Res-$2$} & \multicolumn{3}{c|}{Res-$3$} & \multicolumn{3}{c}{Res-$4$} \\ \cline{2-13} 
 & R-$1$ & R-$5$ & mAP & R-$1$ & R-$5$ & mAP & R-$1$ & R-$5$ & mAP & R-$1$ & R-$5$ & mAP \\ \hline
 
Baseline & $88.36$ & $95.39$ & $71.48$ & $88.36$ & $95.39$ & $71.48$ & $88.36$ & $95.39$ & $71.48$ & $88.36$ & $95.39$ & $71.48$ \\ \hline 

DPB (HP-${1}$) & $88.98$ & $95.22$ & $71.37$ & $90.12$ & $96.08$ & $74.20$ & $89.62$ & $95.67$ & $72.82$ & $89.51$ & $95.55$ & $72.19$ \\

DPB (HP-${2}$) & $90.17$ & $96.35$ & $74.49$ & $90.63$ & $96.67$ & $75.87$ & $90.74$ & $96.39$ & $76.74$ & $90.22$ & $96.34$ & $74.11$ \\

DPB (HP-${5}$) & $90.77$ & $96.44$ & $77.22$ & $91.83$ & $96.89$ & $78.72$ & $91.23$ & $96.74$ & $77.21$ & $90.26$ & $96.29$ & $75.46$ \\

DPB (Latent) & $90.20$ & $96.40$ & $73.28$ & $91.73$ & $96.86$ & $78.48$ & $91.47$ & $96.86$ & $77.80$ & $89.31$ & $96.14$ & $73.71$ \\ 

DPB {(HP-${5}$ + Latent)} & $\bf{91.00}$ & $\bf{96.88}$ & $\bf{76.99}$ & $\bf{92.75}$ & $\bf{97.45}$ & $\bf{80.98}$ & $\bf{91.87}$ & $\bf{97.13}$ & $\bf{78.80}$ & $\bf{91.18}$ & $\bf{97.03}$ & $\bf{78.36}$ \\ \hline

\end{tabular}
\vspace{-0.5cm}
\end{table*}
\vspace{-0.2cm}

\begin{table}[!ht]
\renewcommand\arraystretch{1.0}
\footnotesize
\centering
\caption{\small{
Ablation study of the human-part~(Latent w/o NHP) and non-human part~(Latent w/o HP) in the latent part branch.}}
\vspace{-0.35cm}
\label{table:non-human-latent}
\begin{tabular}{ccc|cc|cc}
\hline
\multirow{2}{*}{HP-5} & Latent & Latent & \multicolumn{2}{c|}{Market Res-$2$} & \multicolumn{2}{c}{Market Res-$3$} \\ \cline{4-7} 
 & w/o NHP & w/o HP & R-$1$ & mAP & R-$1$ & mAP \\ \hline
 
- & - & - & $88.36$ & $71.48$ & $88.36$ & $71.48$ \\ \hline 

 $\times$ & $\checkmark$ & $\times$ & $91.19$ & $77.22$ & $91.12$ & $77.10$ \\
 $\times$ & $\times$ & $\checkmark$ & $91.55$ & $78.25$ & $91.35$ & $77.23$ \\ 
 $\times$ & $\checkmark$ & $\checkmark$ & $\bf{91.73}$ & $\bf{78.48}$ & $\bf{91.47}$ & $\bf{77.80}$ \\ \hline

 & & & \multicolumn{2}{c|}{Market Res-$2$} & \multicolumn{2}{c}{CUHK (detected)} \\ \hline
 $\checkmark$ & $\times$ & $\times$ & $91.83$ & $78.72$ & $67.57$ & $60.02$ \\ 
 $\checkmark$ & $\checkmark$ & $\times$ & $91.97$ & $79.31$ & $68.46$ & $61.98$ \\ 
 $\checkmark$ & $\times$ & $\checkmark$ & $\bf{92.56}$ & $\bf{80.60}$ & $\bf{69.61}$ & $\bf{62.85}$ \\ \hline

\end{tabular}
\vspace{-0.4cm}
\end{table}

\subsection{Ablation study} \label{ablation-study}
\vspace{-0.1cm}

The core idea of DPB lies on the human part branch and the latent part branch. We perform comprehensive ablation studies of them in follows.

\begin{table}[!t]
\renewcommand\arraystretch{1.0}
\footnotesize
\centering
\caption{\small{Comparison of using $1$, $3$ and $5$ DPBs on the Market-$1501$. 
DPB consists of both the human part branch and the latent part branch here.}}
\vspace{-0.35cm}
\label{table:more-block}
\resizebox{\linewidth}{!}
{
\begin{tabular}{l|cc|cccc}
\hline
Method & HP-$5$ & Latent & R-$1$ & R-$5$ & R-$10$ & mAP \\ \hline
Baseline & - & - & $88.36$ & $95.39$ & $97.06$ & $71.48$ \\ \hline

+ $1$ $\times$ DPB & $\checkmark$ & $\times$ & $91.83$ & $96.89$ & $97.95$ & $78.72$ \\
+ $3$ $\times$ DPB & $\checkmark$ & $\times$ & $92.01$ & $97.15$ & $98.16$ & $78.87$ \\
+ $5$ $\times$ DPB & $\checkmark$ & $\times$ & $92.26$ & $97.26$ & $98.20$ & $79.28$ \\ \hline

+ $1$ $\times$ DPB & $\times$ & $\checkmark$  & $91.73$ & $96.86$ & $98.10$ & $78.48$ \\
+ $3$ $\times$ DPB & $\times$ & $\checkmark$  & $92.12$ & $97.32$ & $98.28$ & $80.15$ \\
+ $5$ $\times$ DPB & $\times$ & $\checkmark$  & $92.79$ & $97.65$ & $98.52$ & $80.49$ \\ \hline

+ $1$ $\times$ DPB & $\checkmark$ & $\checkmark$  & $92.75$ & $97.45$ & $98.22$ & $80.98$ \\
+ $3$ $\times$ DPB & $\checkmark$ & $\checkmark$  & $93.28$ & $97.79$ & $98.61$ & $82.08$ \\
+ $5$ $\times$ DPB & $\checkmark$ & $\checkmark$  & $\bf{93.96}$ & $\bf{97.98}$ & $\bf{98.81}$ & $\bf{83.40}$ \\ \hline 
\end{tabular}
}
\vspace{-0.4cm}
\end{table}

\begin{table}[!t]
\renewcommand\arraystretch{1.0}
\footnotesize
\centering
\caption{\small{Comparison of the two branches of DPB on CUHK03.}}
\vspace{-0.35cm}
\label{table:study-cuhk}
\resizebox{\linewidth}{!}
{
\begin{tabular}{l|cc|cccc}
\hline
Method & HP-$5$ & Latent & R-$1$ & R-$5$ & R-$10$ & mAP \\ \hline
Baseline & - & - & $60.29$ & $78.21$ & $84.86$ & $54.79$ \\ \hline
\multirow{3}{*}{+ $5$ $\times$ DPB} & $\checkmark$ & $\times$ & $69.93$ & $83.86$ & $88.90$ & $63.34$ \\
 & $\times$ & $\checkmark$ & $69.84$ & $83.50$ & $89.83$ & $63.25$ \\
 & $\checkmark$ & $\checkmark$ & $\bf{71.55}$ & $\bf{85.71}$ & $\bf{90.80}$ & $\bf{64.23}$ \\ \hline
\end{tabular}
}
\vspace{-0.5cm}
\end{table}

\vspace{-0.55cm}
\paragraph{Influence of the part numbers for human part branch.}
As we can divide the input image into different number of parts in different levels.
we study the impact of the number of different semantic parts (i.e., $K$ = 1, $K$ = 2, $K$ = 5) on the Market-$1501$ benchmark. 
We summarize all of the results in Table \ref{table:res-k}.
The $1^{st}$ row reports the results of baseline model and the the $2^{nd}$ row to $4^{th}$ report the performances that only apply the human part branch with different choices of $K$.
When $K = 1$, there is no extra parsing information added to the network and the performances keep almost the same with the baseline model. 
When $K$ = 2, the human part branch introduces the foreground and the background contextual information to help extract more reliable human context information.
we can observe obvious improvements in R-1 and mAP compared to the previous two results.
The performance improves with larger $K$, which indicates that accurately aggregating contextual information from pixels belonging to same semantic human part is crucial for person re-identification. 
We set $K$ = 5 as the default setting for human part branch if not specified.

\vspace{-0.52cm}
\paragraph{Non-human part in latent part branch.}
The choice of self-attention for latent part branch is mainly inspired by that self-attention can learn to group the similar pixels together without extra supervision (also shown useful in segmentation \cite{OCNet,huang2019isa}). 
Considering that latent part branch is in fact the mixture of the coarse human and non-human part information, we empirically verify that the performance gain from latent part branch is mainly attributed to capturing non-human parts, as shown in Table~\ref{table:non-human-latent}. 
We use binary masks predicted by human parsing model ($K$ = 2) to control the influence of human or non-human regions within latent part branch.
Here we study two kinds of settings: 
(1) only use non-human part information within latent part branch. We apply binary human masks ($1$ for non-human pixels and $0$ for human pixels) to remove the influence of pixels predicted as human parts, which is called as Latent w/o HP.
(2) only use human part information within latent part branch. We also apply binary human masks ($1$ for human pixels and $0$ for non-human pixels) to remove the influence of pixels predicted as non-human parts, which is called as Latent w/o NHP. 
It can be seen that the gain of latent part branch mainly comes from the help of non-human part information, Latent w/o HP outperforms Latent w/o NHP and is very close to the original latent part branch.

\begin{figure}[tb]
\centering
\includegraphics[width=0.48\textwidth]{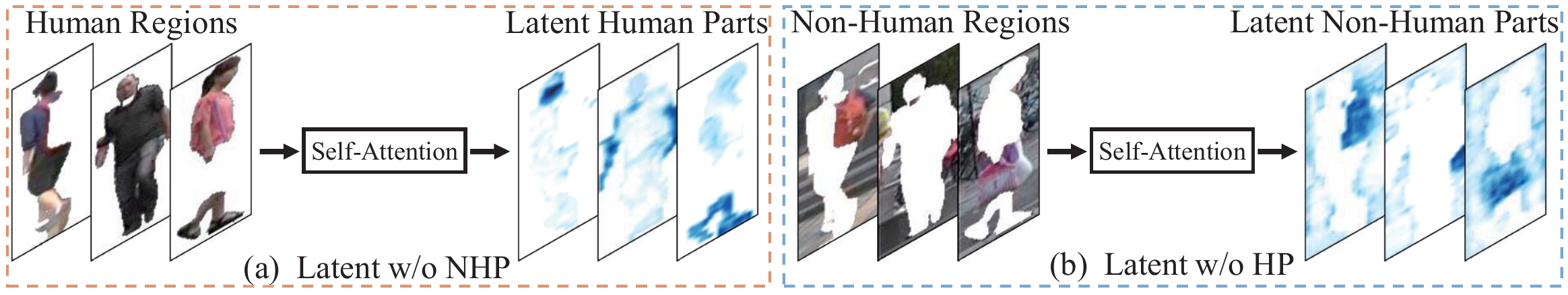}
\vspace{-0.67cm}
\caption{\small{Latent w/o NHP vs. Latent w/o HP: \emph{Latent w/o NHP} only applies self-attention on the human part regions while \emph{Latent w/o HP} only applies self-attention on the non-human part regions.
The human/non-human part regions are
based on the human parsing prediction.}}
\label{fig:latent}
\vspace{-0.65cm}
\end{figure}

Besides, we study the contribution of latent branch when applying human part branch (HP-5).
We choose DPB (HP-5) inserted after Res-$2$ as our baseline and add latent part branch that applies self-attention on either the human regions only (Latent w/o NHP in Figure~\ref{fig:latent}) or non-human regions only (Latent w/o HP in Figure~\ref{fig:latent}). 
It can be seen that DPB (HP-5 + Latent w/o HP) largely outperforms DPB (HP-5 + Latent w/o NHP) and is close to DPB (HP-5 + Latent), which further verifies the effectiveness of latent part branch is mainly attributed to exploiting the non-human parts.

\vspace{-0.55cm}
\paragraph{Complementarity of two branches.}
Dual part-aligned block (DPB) consists of two branches: human part branch and latent part branch. The human part branch helps improve the performance by eliminating the influence of noisy background context information, 
and the latent part branch introduces latent part masks to surrogate various non-human parts.

We empirically show that the two branches are complementary through the experimental results on the $6^{th}$ row of Table~\ref{table:res-k}. 
It can be seen that combining both the human part-aligned representation and the latent part-aligned representation boosts the performance for all stages. 
We can draw the following conclusions from Table~\ref{table:more-block} and Table~\ref{table:study-cuhk}: (i) Although the latent part masks are learned from scratch, DPB (latent) achieves comparable results with the human part branch in general, which carries stronger prior information of human parts knowledge, showing the importance of the non-human part context. (ii) Human part branch and latent part branch are complementary to each other. In comparison to the results only using a single branch, inserting $5 \times$ DPB attains $1\%$ and $3\%$ gain in terms of R-1 and mAP on Market-1501, $1.6\%$ and $1\%$ gain in terms of R-1 and mAP on CUHK03, respectively.

\begin{figure}[!tb]
\hspace{-0.36cm}
\includegraphics[scale=0.38]{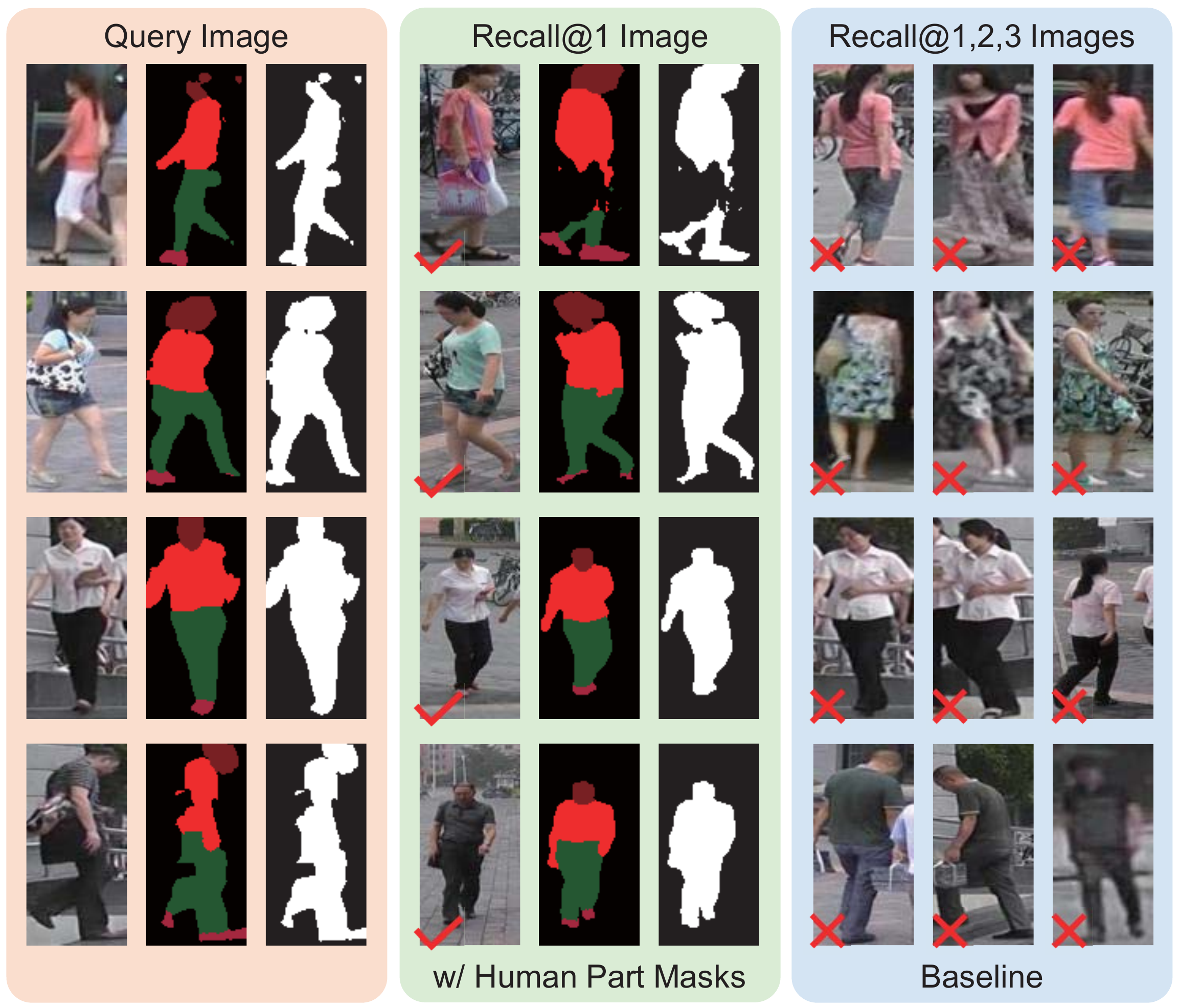}
\vspace{-0.6cm}
\caption{
\small{Comparison of Baseline and $P^2$-Net that only employs the human part branch.
For all of the four query images, Recall@3 of the Baseline method is 0, while the Recall@1 of the $P^2$-Net (w/ Human Part Masks) is 1.
The $1^{st}$ and $2^{nd}$ rows illustrate the cases that the bag is visible in one viewpoint but invisible in other viewpoints, 
human part masks eliminate the influence of the bags as the bags are categorized as background.
The $3^{rd}$ and $4^{th}$ rows illustrate cases that the area of person only
occupies small proportions of the whole images and the background context information
leads to poor performance, human part masks can eliminate the influence of background regions.
}}
\label{fig:DPR_hard_parsing}
\vspace{-0.5cm}
\end{figure}

\begin{figure}[!tb]
\hspace{-0.31cm}
\includegraphics[scale=0.38]{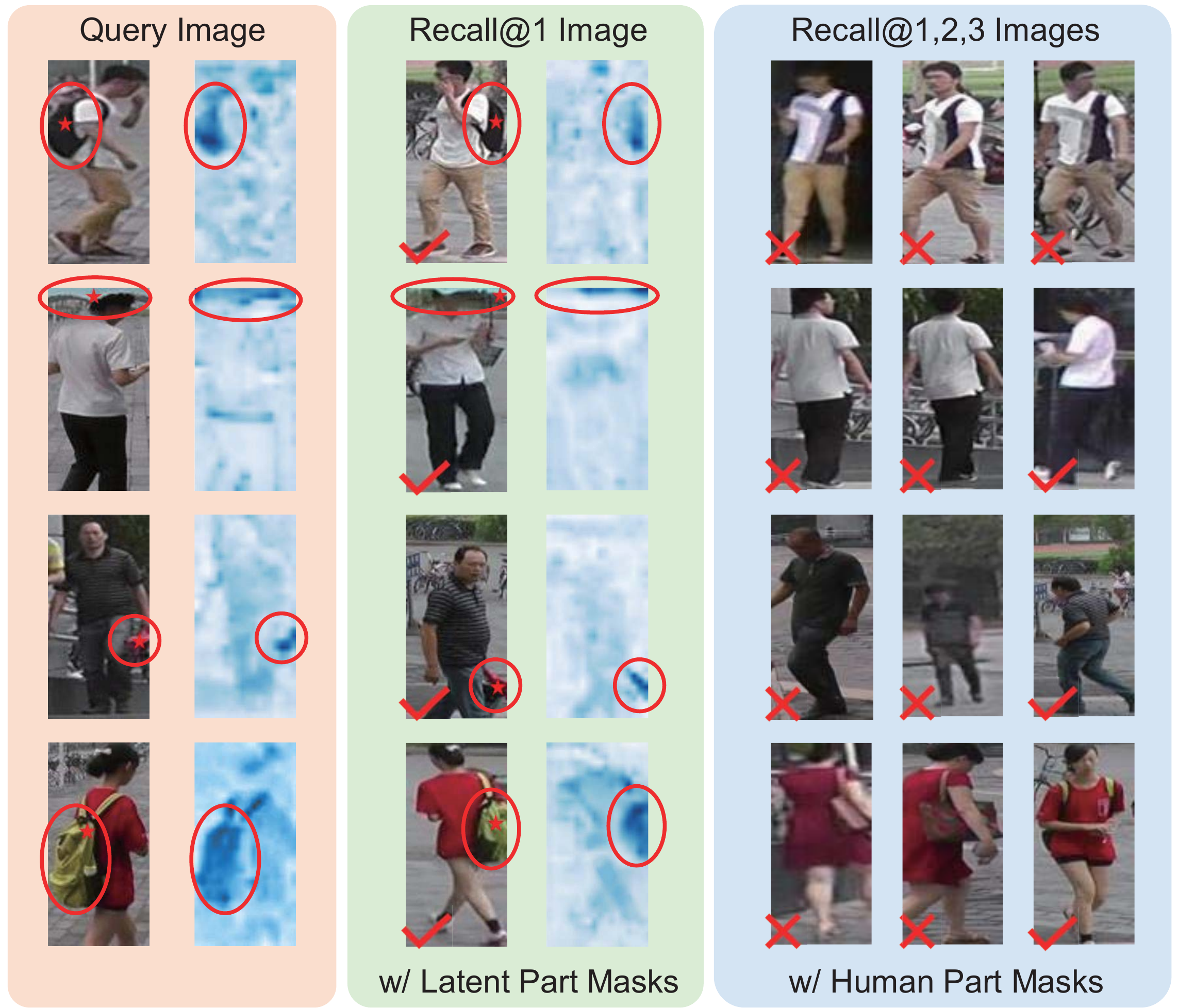}
\vspace{-0.5cm}
\caption{
\small{Comparison of $P^2$-Net (w/ Latent Part Masks) and $P^2$-Net (w/ Human Part Masks).
There exist some important non-human parts in all of the four query images.
The $P^2$-Net (w/ Human Part Masks) categorizes these crucial non-human parts as background and fails to return the correct image at Recall@1.
The $P^2$-Net (w/ Latent Part Masks) predicts the latent part mask associated with these non-human
parts, which successfully returns the correct image at Recall@1.
It can be seen that the predicted latent part masks serves as reliable surrogate for the non-human part.
}}
\label{fig:DPR_soft_parsing}
\vspace{-0.5cm}
\end{figure}

\vspace{-0.03cm}
We visualize the predicted human part masks to illustrate how it helps improve the performance in Figure~\ref{fig:DPR_hard_parsing}.
For all query images above,
baseline method fails to return the correct images of the same identity while
we can find out the correct images by employing human part masks.
In summary,
we can see that the context information of the non-informative background influences the final results and the human part masks eliminate the influence of these noisy context information.

\vspace{-0.03cm}
There also exist large amounts of scenarios that non-human part context information is the key factor. 
We illustrate some typical examples in Figure~\ref{fig:DPR_soft_parsing},
and we mark the non-human but informative parts with red circles. 
For example, the $1^{st}$ and $4^{th}$ row illustrate that 
mis-classifying the \textbf{bag} as background causes the failure of the human part masks based method.
Our approach addresses these failed cases through learning the latent part masks and it can be seen that the predicted latent part masks within latent part branch well surrogate the non-human but informative parts. 
In summary, the human part branch benefits from the latent part branch through dealing with crucial non-human part information.

\vspace{-0.55cm}
\paragraph{Number of DPB.}
To study the influence of the numbers of DPB (with human part representation only, with latent part representation only and with both human and latent part representations), we add 1 block (to Res-2), 3 blocks (2 to Res-2, and 1 to Res-3) and 5 blocks (2 to Res-2, and 3 to Res-3) within the backbone network. 
As shown in Table~\ref{table:more-block}, 
more DPB blocks lead to better performance. 
We achieve the best performance with 5 DPB blocks, 
which boosts the R-1 accuracy and mAP by $5.6\%$ and $11.9\%$, respectively.
We set the number of DPB block as $5$ in all of our state-of-the-art experiments.

\vspace{-0.3cm}
\subsection{Comparison with state-of-the-art}
We empirically verify the effectiveness of our approach with a series of state-of-the-art (SOTA) results on all of the three benchmarks.
We illustrate more details as following.

\begin{table}[!t]
\renewcommand\arraystretch{1.0}
\centering
\small
\caption{\small{Comparison with the SOTA on Market-$1501$.}}
\vspace{-0.35cm}
\label{table:market}
\begin{tabular}{l|cccc}
\hline
Method & R-$1$ & R-$5$ & R-$10$ & mAP \\ 
\hline
\small Spindle \cite{spindlenet} & $76.9$ & $91.5$ & $94.6$ & - \\
\small MGCAM \cite{mgcam} & $83.8$ & - & - & $74.3$ \\
\small PDC \cite{pdc} & $84.1$ & $92.7$ & $94.9$ & $63.4$ \\
\small AACN \cite{aacn} & $85.9$ & - & - & $66.9$ \\ 
\small PSE \cite{pse} & $87.7$ & $94.5$ & $96.8$ & $69.0$ \\ 
\small PABR \cite{jingdongeccv18} & $90.2$ & $96.1$ & $97.4$ & $76.0$ \\
\small SPReID \cite{spreid} & $92.5$ & $97.2$ & $98.1$ & $81.3$ \\ 
\small MSCAN \cite{li2017learning} & $80.3$ & - & - & $57.5$ \\
\small DLPAR \cite{zhaoliming} & $81.0$ & $92.0$ & $94.7$ & $63.4$ \\
\small SVDNet \cite{svdnet} & $82.3$ & $92.3$ & $95.2$ & $62.1$ \\
\small DaF \cite{daf} & $82.3$ & - & - & $72.4$ \\
\small JLML \cite{jlml} & $85.1$ & - & - & $65.5$ \\
\small DPFL \cite{chen2018person} & $88.9$ & - & - & $73.1$ \\
\small HA-CNN \cite{hacnn} & $91.2$ & - & - & $75.7$ \\
\small SGGNN \cite{sggnn} & $92.3$ & $96.1$ & $97.4$ & $\underline{82.8}$ \\
\small GSRW \cite{randomwalk} & $92.7$ & $96.9$ & $98.1$ & $82.5$ \\ 
\small PCB + RPP \cite{pcb} & $\underline{93.8}$ & $\underline{97.5}$ & $\underline{98.5}$ & $81.6$ \\ \hline
\small $P^2$-Net & $94.0$ & $98.0$ & $98.8$ & $83.4$ \\
\small $P^2$-Net (+ triplet loss) & $\bf{95.2}$ & $\bf{98.2}$ & $\bf{99.1}$ & $\bf{85.6}$ \\ \hline
\end{tabular}
\vspace{-0.3cm}
\end{table}

We illustrate the comparisons of our $P^2$-Net with the previous state-of-the-art methods on Market-1501 in Table~\ref{table:market}. 
Our $P^2$-Net outperforms all the previous methods by a large margin.
We achieve a new SOTA performance such as R-1=$95.2\%$ and mAP=$85.6\%$ respectively. 
Especially,
our $P^2$-Net outperforms the previous PCB by 1.8\% in mAP without using multiple softmax losses for training. 
When equiped with the triplet loss, our $P^2$-Net still outperforms the PCB by 1.4\% and 4.0\% in terms of R-1 and mAP, respectively.
Besides, our proposed $P^2$-Net also outperforms the SPReID \cite{spreid} by 2.7\% measured by R-1 accuracy.

We summarize the comparisons on DukeMTMC-reID in Table~\ref{table:duke}.
It can be seen that $P^2$-Net surpasses all previous SOTA methods. SPReID \cite{spreid} is the method has the closest performance with us in R-1 accuracy.
Notably, SPReID train their model with more than $10$ extra datasets
to improve the performance while we only use the original dataset as training set.

\begin{table}[!t]
\renewcommand\arraystretch{1.0}
\centering
\small
\caption{\small{Comparison with the SOTA on DukeMTMC-reID.}}
\vspace{-0.35cm}
\label{table:duke}
\begin{tabular}{l|cccc}
\hline
Method & R-$1$ & R-$5$ & R-$10$ & mAP \\ \hline 
\small AACN \cite{aacn} & $76.8$ & - & - & $59.3$ \\
\small PSE \cite{pse} & $79.8$ & $89.7$ & $92.2$ & $62.0$ \\ 
\small PABR \cite{jingdongeccv18} & $82.1$ & $90.2$ & $92.7$ & $64.2$ \\
\small SPReID \cite{spreid} & $\underline{84.4}$ & $\underline{91.9}$ & $\underline{93.7}$ & $\underline{71.0}$ \\ 
\small SBAL \cite{sbal} & $71.3$ & - & - & $52.4$ \\
\small ACRN \cite{schumann2017person} & $72.6$ & $84.8$ & $88.9$ & $52.0$ \\
\small SVDNet \cite{svdnet} & $76.7$ & $86.4$ & $89.9$ & $56.8$ \\
\small DPFL \cite{chen2018person} & $79.2$ & - & - & $60.6$ \\ 
\small SVDEra \cite{reranking} & $79.3$ & - & - & $62.4$ \\
\small HA-CNN \cite{hacnn} & $80.5$ & - & - & $63.8$ \\
\small GSRW \cite{randomwalk} & $80.7$ & $88.5$ & $90.8$ & $66.4$ \\
\small SGGNN \cite{sggnn} & $81.1$ & $88.4$ & $91.2$ & $68.2$ \\
\small PCB + RPP \cite{pcb} & $83.3$ & $90.5$ & $92.5$ & $69.2$ \\ \hline
\small $P^2$-Net & $84.9$ & $92.1$ & $94.5$ & $70.8$ \\
\small $P^2$-Net (+ triplet loss) & $\bf{86.5}$ & $\bf{93.1}$ & $\bf{95.0}$ & $\bf{73.1}$ \\
\hline
\end{tabular}
\vspace{-0.3cm}
\end{table}

\begin{table}[!t]
\renewcommand\arraystretch{1.0}
\centering
\small
\caption{\small{Comparison with the SOTA on CUHK03.}}
\vspace{-0.35cm}
\label{table:cuhk}
\begin{tabular}{l|cc|cc}
\hline
\multirow{2}{*}{Method} & \multicolumn{2}{c|}{labeled} & \multicolumn{2}{c}{detected} \\
\cline{2-5}
~ & R-$1$ & mAP & R-$1$ & mAP \\ \hline 
\small DaF \cite{daf} & $27.5$ & $31.5$ & $26.4$ & $30.0$ \\
{\small SVDNet \cite{svdnet}} & $40.9$ & $37.8$ & $41.5$ & $37.3$ \\
{\small DPFL \cite{chen2018person}} & $43.0$ & $40.5$ & $40.7$ & $37.0$ \\
\small HA-CNN \cite{hacnn} & $44.4$ & $41.0$ & $41.7$ & $38.6$ \\
\small SVDEra \cite{reranking} & $49.4$ & $45.1$ & $48.7$ & $43.5$ \\
{\small MGCAM \cite{mgcam}}& $\underline{50.1}$ & $\underline{50.2}$ & $46.7$ & $46.9$ \\
{\small PCB + RPP \cite{pcb}} & - & - & $\underline{63.7}$ & $\underline{57.5}$ \\ \hline
{\small $P^2$-Net} & 75.8 & 69.2 & 71.6 & 64.2 \\ 
{\small $P^2$-Net (+ triplet loss)} & $\bf{78.3}$ & $\bf{73.6}$ & $\textbf{74.9}$ & $\bf{68.9}$ \\ \hline
\end{tabular}
\vspace{-0.5cm}
\end{table}

Last, we evaluate our $P^2$-Net on CUHK03 dataset.
We follow the training/testing protocol proposed by \cite{reranking}. 
As illustrated in Table~\ref{table:cuhk},
our $P^2$-Net outperforms previous SOTA method MGCAM \cite{mgcam} by $28.2\%$ measured by R-1 accuracy and $23.4\%$ measured by mAP.
For the CUHK03-detected dataset, our $P^2$-Net still outperforms previous SOTA method PCB+RPP \cite{pcb} by $11.2\%$ measured by R-1 accuracy and $11.4\%$ measured by mAP.

In conclusion, our $P^2$-Net outperforms all the previous approaches by a large margin and achieves new state-of-the-art performances on all the three challenging person re-identification benchmarks.

\vspace{-0.1cm}
\section{Conclusion}
\vspace{-0.1cm}
In this work, we propose a novel \textbf{dual part-aligned representation}
scheme to address the non-human part misalignment problem for person re-identification.
It consists of a human part branch
and a latent part branch to tackle both 
human part misalignment and
non-human part misalignment problem. 
The human part branch adopts off-the-shelf human parsing model to inject structural prior information by capturing the predefined semantic human parts for a person. 
The latent part branch adopts a self-attention mechanism to help capture the detailed part categories beyond the injected prior information. 
Based on our dual part-aligned representation, 
we achieve new state-of-the-art performances on all of the three benchmarks including Market-1501, DukeMTMC-reID and CUHK03.

\vspace{-0.1cm}
\section*{Acknowledgement}
\vspace{-0.2cm}
This work was supported by the National Natural Science Foundation of China under Grant 61671027 and the National Key Basic Research Program of China under Grant 2015CB352303.

{\small
\bibliographystyle{ieee_fullname}
\bibliography{egbib}
}

\newpage
\section{Appendix}

In this supplementary material, we provide the complexity analysis of our method, details about triplet loss, and show more typical experiment results and cases on  DukeMTMC-ReID and CUHK03.

\subsection{Triplet loss}
As mentioned in Sec 3.2, we use triplet loss to improve the performance in final results. The details are as folows:
(1) We prepare each mini-batch by randomly sampling $16$ classes (identities) and $4$ images for each class.
(2) We set the weight rate as 1:1 on all three datasets.
(3) Given a mini-batch of $64$ samples, we construct a triplet for each sample by choosing the hardest positive sample and the hardest negative sample measured by their Euclidean distances.

\subsection{Strategies for inserting DPB.}
We do the ablation study to find the results of adding DPB after different Res-$k$ residual blocks. As shown in the Table~\textcolor{red}{1} of main paper, we can find that all types of blocks (DPB/Human Part Branch/Latent Part Branch) achieve better performances when they are inserted after the Res-2 and Res-3 stages, compared to Res-1 and Res-4 stages. Specifically, the DPB improves the Rank-1 accuracy and mAP by 4.4\% and 9.5\% when inserted after res-2 stage, 3.4\% and 7.3\% when inserted after res-3 stage, respectively. One possible explanation is that the feature map from Res-1 has more precise localization information but less semantic information, and the deeper feature map from Res-4 is insufficient to provide precise spatial information. In conclusion, Res-2 and Res-3 can benefit more from the proposed DPB. So the 5$\times$DPB in all experiments means that we add 2 DPB blocks to Res-$2$ and 3 DPB blocks to Res-$3$, if not specified.

\subsection{Complexity analysis}
We compare the proposed model with ResNet-50 and ResNet-101 in model size and computation complexity, measured by the number of parameers and FLOPs during inference on CUHK$03$. And we test the inference time of each forward pass on a single GTX 1080Ti GPU with CUDA8.0 given an input image of size $3\times384\times128$. Table~\ref{table:depth-compare} shows that our method outperforms ResNet-101 with smaller model size, less computation amount and faster inference speed, the improvement of $P^2$-Net is not just because the added depth to the baseline model.

\subsection{Experiments on DukeMTMC-reID}
To further verify that the latent part branch and the human part branch are complementary, we also conduct the controlled experiments on both DukeMTMC-ReID and CUHK03.

We present the results on DukeMTMC-reID in Table~\ref{table:block-duke}.
It can be seen that DPB achieves better performance than either only employing the latnet
part branch or only employing the human part branch. 
e.g., 
``$1$ $\times$ DPB" improves the mAP of ``$1$ $\times$ DPB (HP-5)" from 
$66.99$ to $67.93$. 
``$5$ $\times$ DPB" improves the mAP of ``$5$ $\times$ HPP (HP-5)" from 
$68.64$ to $70.84$.

\begin{table}[!t]
\renewcommand\arraystretch{1.0}
\centering
\footnotesize
\caption{\small{Complexity comparison of DPB/Baseline on CUHK03.}}
\begin{tabular}{l|c|c|c|c|c|c} \hline
Method & $5\times$DPB &  Params & FLOPs & Time & R-$1$ & mAP \\
\hline
R-$50$ & $\times$      & $24.2$M & $14.9$G & 19ms & $60.29$ & $54.79$ \\
R-$101$ & $\times$     & $43.2$M & $22.1$G & 32ms & $68.14$ & $63.45$ \\
R-$50$ & $\checkmark$  & $31.6$M & $18.6$G & 27ms & $\bf{71.55}$ & $\bf{64.23}$ \\
\hline
\end{tabular}
\label{table:depth-compare}
\end{table}

\begin{figure*}[!htb]
\centering
\includegraphics[width=\textwidth]{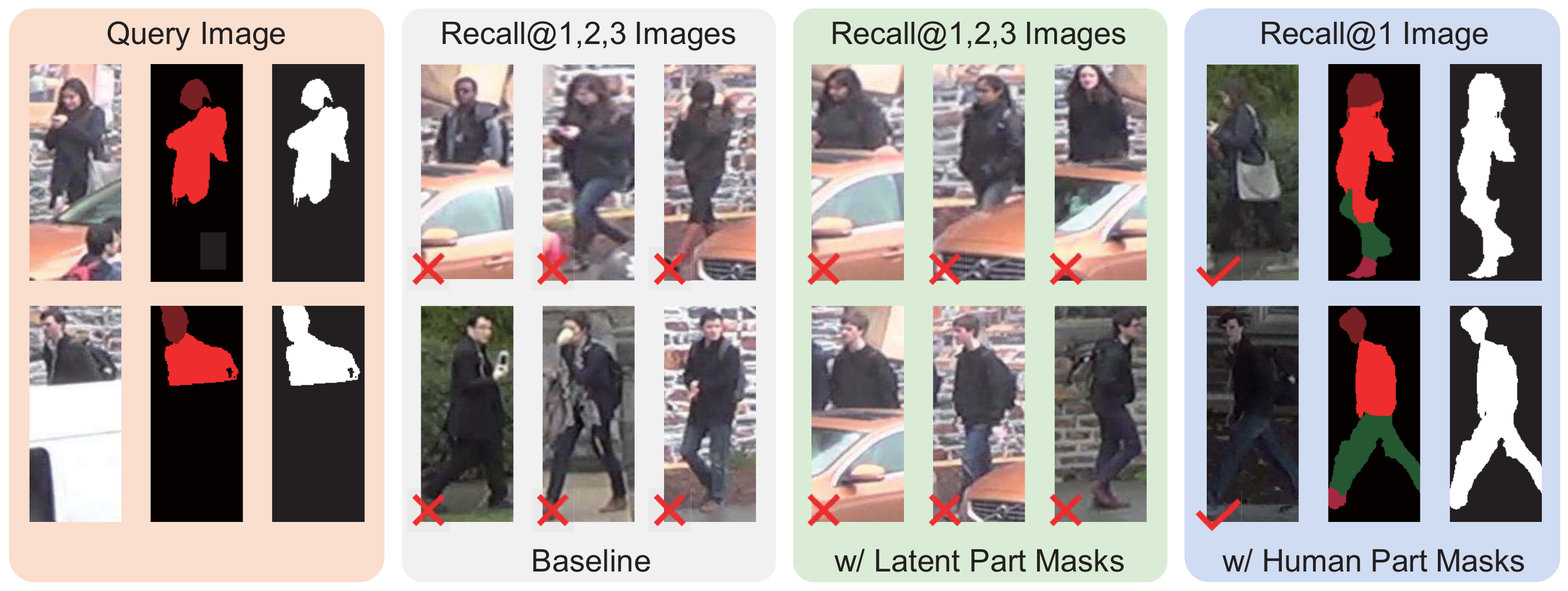}
\caption{Comparison of Baseline, DPB (w/ Latent Part Masks) and DPB (w/ Human Part Masks) on \textbf{DukeMTMC-ReID}. We denote $P^2$-Net that only employs human part branch as the method w/ Human Part Masks. 
Both these two query images suffer from the problem 
of occlusions and contain useless or misleading background
information.
Both the baseline and DPB (w/ Latent Part Masks) fail to
return the correct results within the top 3 positions while
DPB (w/ Human Part Masks) returns the correct result at top 1 position.
}
\label{fig:duke_parsing}
\end{figure*}

We present the advantages of human part branch in
Figure~\ref{fig:duke_parsing} .
The results with human part branch 
perform more robust compared with the 
results of baseline and the results
with the latent part branch.
For example, the query image on the $1^{st}$
line carries the misleading information 
caused by the part of a car.
Both the baseline method and the method with latent part branch return the images
carrying parts of the car,
and the method with human part branch
returns the correct result by
removing the influence of the car.

\begin{table}[!t]
\renewcommand\arraystretch{1.0}
\centering
\caption{\small{Comparison experiments on \textbf{DukeMTMC-ReID}. 
DPB (HP-5) only uses the human part branch and sets $K=5$.
DPB (Latent) only uses the latent part branch.
DPB uses both the human part branch and the latent part branch.}}
\label{table:block-duke}
\begin{tabular}{l|cccc}
\hline
Method & R-$1$ & R-$5$ & R-$10$ & mAP \\ \hline
Baseline & $79.85$ & $89.81$ & $92.19$ & $62.57$ \\ \hline

$1$ $\times$ DPB (HP-5) & $83.04$ & $91.18$ & $93.22$ & $66.99$ \\
$1$ $\times$ DPB (Latent) & $82.20$ & $90.33$ & $92.69$ & $65.09$ \\
$1$ $\times$ DPB & $\bf{83.80}$ & $\bf{91.38}$ & $\bf{93.58}$ & $\bf{67.93}$ \\ \hline

$5$ $\times$ DPB (HP-5) & $84.08$ & $91.82$ & $94.10$ & $68.64$ \\ 
$5$ $\times$ DPB (Latent) & $84.45$ & $91.97$ & $94.25$ & $69.07$ \\ 
$5$ $\times$ DPB & $\bf{84.91}$ & $\bf{92.08}$ & $\bf{94.45}$ & $\bf{70.84}$ \\ \hline 
\end{tabular}
\end{table}

\begin{table}[!t]
\renewcommand\arraystretch{1.0}
\centering
\caption{\small{Comparison experiments on \textbf{CUHK03}. 
DPB (HP-5) only uses the human part branch and sets $K=5$.
DPB (Latent) only uses the latent part branch.
DPB uses both the human part branch and the latent part branch.}} 
\label{table:block-cuhk}
\begin{tabular}{l|cccc}
\hline
Method & R-$1$ & R-$5$ & R-$10$ & mAP \\ \hline
Baseline & $60.29$ & $78.21$ & $84.86$ & $54.79$ \\ \hline
$1$ $\times$ DPB (HP-5) & $67.57$ & $81.32$ & $87.36$ & $60.02$ \\
$1$ $\times$ DPB (Latent) & $68.59$ & $83.14$ & $87.96$ & $61.75$ \\
$1$ $\times$ DPB & $\bf{70.43}$ & $\bf{84.50}$ & $\bf{89.64}$ & $\bf{63.93}$ \\ \hline
$5$ $\times$ DPB (HP-5) & $69.93$ & $83.86$ & $88.90$ & $63.34$ \\ 
$5$ $\times$ DPB (Latent) & $69.84$ & $83.50$ & $89.83$ & $63.25$ \\ 
$5$ $\times$ DPB & $\bf{71.55}$ & $\bf{85.71}$ & $\bf{90.80}$ & $\bf{64.23}$ \\ \hline 
\end{tabular}
\end{table}
 
\begin{figure*}[!htb]
\centering
\includegraphics[width=\textwidth]{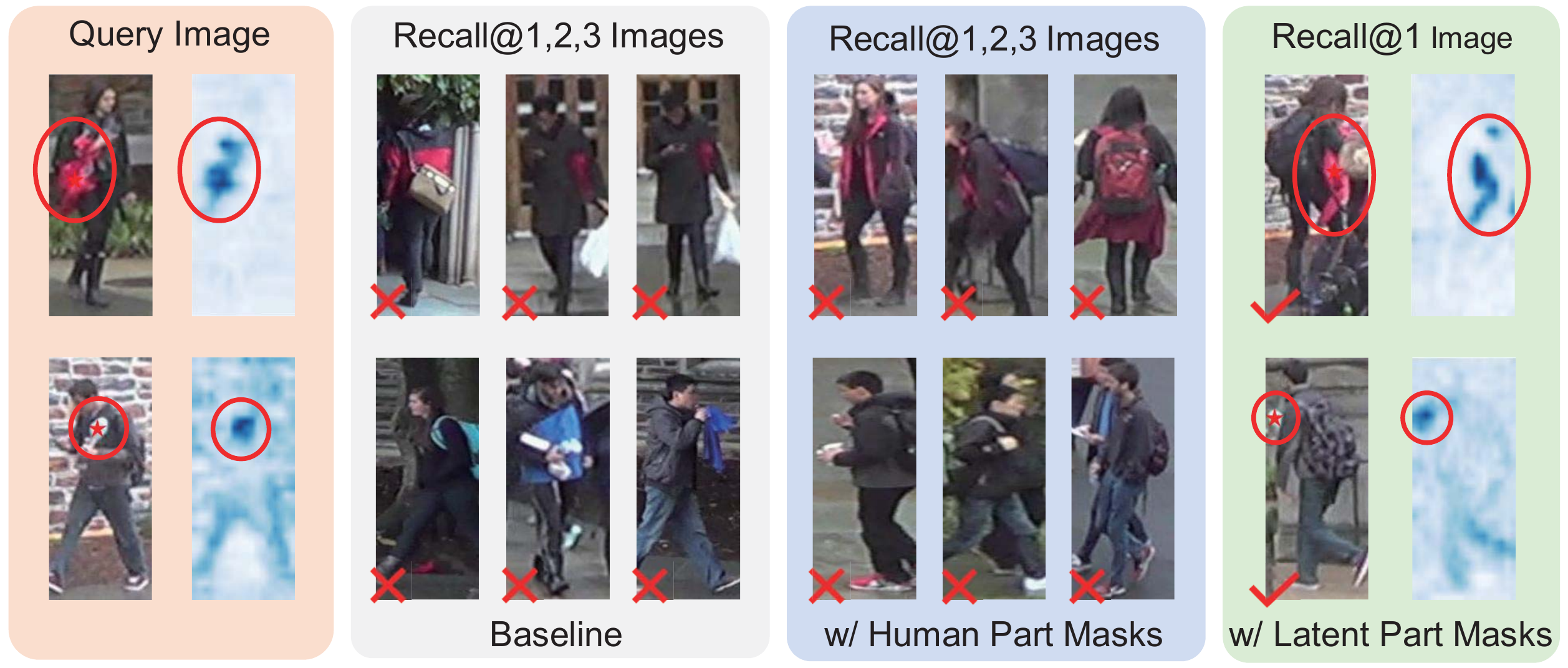}
\caption{Comparison of Baseline, DPB (w/ Human Part Masks) and DPB (w/ Latent Part Masks) on \textbf{DukeMTMC-ReID}. There exist some important non-human parts within all these two query images. The DPB (w/ Human Part Masks) categorizes these important parts to background and fails to return the correct image. The DPB (w/ Latent Part Masks) predicts the latent part mask associated with these parts, which helps to find the correct image.
}
\label{fig:duke_attn}
\end{figure*}

We also present the benefits of latent part branch 
in Figure~\ref{fig:duke_attn}.
The failed cases in both the baseline and the method with human part branch are solved by using the latent part masks generated by latent part branch.
It can be seen that these latent part masks 
capture some non-human but important part information that fail to be captured by both the baseline method and the method with only human part branch.
We can conclude that latent part branch and human part branch are complementary accordingly.

\subsection{Experiments on CUHK03}
We report the results on CUHK03 (detected) in Table~\ref{table:block-cuhk}.
And we can find similar performance improvements by combining 
the human part branch and the latent part branch. 
e.g., 
``$1$ $\times$ DPB" improves the mAP of ``$1$ $\times$ DPB (HP-5)" from 
$60.02$ to $63.93$.
``$5$ $\times$ DPB" improves the mAP of ``$5$ $\times$ DPB (HP-5)" from 
$63.34$ to $64.23$.

Our approach boosts the performance of baseline model by a large margin, especially on CUHK03 dataset, the probable reasons are (i) the quality is better (less blurring effects, higher image
resolutions: $266\times90$ in CUHK03, $128\times64$ in Market-1501), thus the DBP can estimate 
more accurate human parsing and latent attention results.
(ii) the background across different images is more noisy, DPB can remove the influence of the background.

\end{document}